\documentclass[onecolumn]{article}
\usepackage{helvet}

\usepackage[a4paper,margin=0.5in]{geometry}
\usepackage{cite}
\usepackage{amsmath,amssymb,amsfonts}
\usepackage{graphicx}
\usepackage{textcomp}
\usepackage{hyperref}
\usepackage{siunitx}
\usepackage{makecell}
\usepackage{booktabs}
\usepackage{tabularx}  

\usepackage[caption=false]{subfig}
\usepackage{algorithm}
\usepackage{algpseudocode}
\usepackage{titlesec}  
\titleformat{\section}{\large\bfseries}{\thesection.}{0.5em}{}
\titleformat{\subsection}{\normalsize\bfseries}{\thesubsection.}{0.5em}{}
\usepackage{authblk}   
\usepackage{float}
\usepackage{nth}
\makeatletter
\def\fps@figure{tp}
\def\fps@table{tp}
\makeatother

\begin{document}
\title{Federated nnU-Net for Privacy-Preserving Medical Image Segmentation}

\author[*1]{Grzegorz Skorupko}
\author[*1]{Fotios Avgoustidis}
\author[1]{Carlos Martín-Isla}
\author[1]{Lidia Garrucho}
\author[1]{Dimitri A. Kessler}
\author[1]{Esmeralda Ruiz Pujadas}
\author[1]{Oliver Díaz}
\author[2]{Maciej Bobowicz}
\author[2]{Katarzyna Gwoździewicz}
\author[3]{Xavier Bargalló}
\author[4]{Paulius Jaruševičius}
\author[1]{Richard Osuala}
\author[1]{Kaisar Kushibar}
\author[1,5]{Karim Lekadir}

\affil[1]{Artificial Intelligence in Medicine Laboratory (BCN-AIM), Departament de Matemàtiques i Informàtica, Universitat de Barcelona, 08007 Barcelona, Spain}
\affil[2]{Medical University of Gdańsk (GUMed), 80-210 Gdańsk, Poland}
\affil[3]{Hospital Clínic de Barcelona (HCB), 08036 Barcelona, Spain}
\affil[4]{Lithuanian University of Health Sciences, 44307 Kaunas, Lithuania}
\affil[5]{Institució Catalana de Recerca i Estudis Avançats (ICREA), 08010 Barcelona, Spain}



\begingroup
\renewcommand{\thefootnote}{}
\footnotetext[2]{* These authors contributed equally to this work.}
\footnotetext[3]{Corresponding author: Grzegorz Skorupko (\texttt{grzegorz.skorupko@ub.edu}).}
\endgroup

\maketitle
\begin{abstract}
\textbf{The nnU-Net framework has played a crucial role in medical image segmentation and has become the gold standard in multitudes of applications targeting different diseases, organs, and modalities. However, so far it has been used primarily in a centralized approach where the collected data is stored in the same location where nnU-Net is trained. This centralized approach has various limitations, such as potential leakage of sensitive patient information and violation of patient privacy. Federated learning has emerged as a key approach for training segmentation models in a decentralized manner, enabling collaborative development while prioritising patient privacy. In this paper, we propose FednnU-Net, a plug-and-play, federated learning extension of the nnU-Net framework. To this end, we contribute two federated methodologies to unlock decentralized training of nnU-Net, namely, Federated Fingerprint Extraction (FFE) and Asymmetric Federated Averaging (AsymFedAvg). We conduct a comprehensive set of experiments demonstrating high and consistent performance of our methods for breast, cardiac and fetal segmentation based on a multi-modal collection of 6 datasets representing samples from 18 different institutions. To democratize research as well as real-world deployments of decentralized training in clinical centres, we publicly share our framework at \textit{https://github.com/faildeny/FednnUNet}.}
\end{abstract}

data privacy, federated learning, image segmentation, nnU-Net, nnUNet
\bigskip
\section*{Introduction}
\label{sec:introduction}

Medical image segmentation plays a key role in clinical workflows to accurately delineate anatomical structures and help with diagnosis, treatment planning, and disease monitoring \cite{azad2024medical,xu_2022_a,Bernard2018,hollon2020near,esmaeili2018direction}. Deep learning methods, especially convolutional neural networks (CNNs)~\cite{LeCun2015,Fukushima1980}, have shown great success in this field. One of the most effective approaches is the no-new-U-Net framework (nnU-Net)~\cite{Isensee2021} which, over the years, has established itself as the go-to framework for medical image segmentation task, evidenced by its multiple first-place finishes on public leaderboards and its widespread adoption as the foundation for further model development across diverse segmentation challenges~\cite{nnUNet_as_basis, Ise_nnUNet_MICCAI2024}.

nnU-Net extends the widely adopted U-Net architecture \cite{ronneberger2015u}, which utilizes an encoder-decoder structure with skip connections to capture both semantic and localization information. Going beyond the original U-Net, nnU-Net automatically configures network architectures tailored to a given dataset, provides a robust training schedule and data augmentation strategy, as well as postprocessing methods to further improve the final performance. This “no-new” (nn) principle largely eliminates the manual effort associated with hyperparameter tuning and architectural engineering, rendering nnU-Net both powerful and highly convenient across a broad range of biomedical tasks \cite{Isensee2021,Liang2024,choi2024deep, nnunet_vs_other_unets}.

Despite these advancements, the development of robust segmentation models often requires aggregating extensive datasets from multiple institutions to capture diverse patient populations and imaging protocols. However, the consolidation of such data is frequently impeded by stringent data privacy regulations, ethical considerations, and logistical challenges associated with data sharing. Federated learning (FL)~\cite{McMahan2017} was introduced as a promising solution to these challenges by facilitating collaborative model training across multiple institutions without the need to exchange raw data.

Although nnU-Net has become a standard for medical image segmentation in centralized settings, its use in FL is challenging due to its dataset-specific customization which involves preprocessing, normalization, model choice, and training schedules based on the underlying data~\cite{Isensee2021}. The model architecture itself includes design decisions regarding the size and stride of convolutional kernels, as well as the number of network blocks. Such level of customization fundamentally conflicts with standard FL methods like FedAvg~\cite{McMahan2017} that require uniform model architectures and training procedures across clients.

To solve these limitations, we introduce \textbf{FednnU-Net}--the first fully federated privacy-preserving implementation of nnU-Net. FednnU-Net is specifically designed to maintain the automated and out-of-the-box nature of nnU-Net while overcoming the challenges associated with decentralized, heterogeneous data.

To this end, we propose two methods---\textbf{Federated Fingerprint Extraction (FFE)} and \textbf{Asymmetric Federated Averaging (AsymFedAvg)} to unlock distributed training of nnU-Net. These methods address the challenges posed by variations in dataset properties across different centers, ensuring that each institution maximizes the benefit from shared knowledge training. Our experiments demonstrate that both methods consistently outperform local, non-collaborative training setups for 2D and 3D segmentation tasks across six public and non-public datasets. 

By extending nnU-Net’s automated pipeline to FL and introducing methods to handle data heterogeneity, this work advances the development of robust, scalable, and privacy-preserving medical image segmentation solutions. Furthermore, its modular design also unlocks access to bringing a wide range of nnU-Net extensions to the federated settings, including improved training efficiency through e.g. curriculum learning~\cite{fischer2024progressive}, enhanced model explainability~\cite{wang2024towards}, continual learning~\cite{gonzalez2023lifelong}, or uncertainty estimation~\cite{zhao2022efficient, joshi2024leveraging, vorberg2024bayesian}.

\section*{Related Work}\label{sec:related_work}

\subsection*{Federated learning in medical imaging}

FL has been widely adopted in medical imaging for tasks such as disease classification, lesion detection, and organ segmentation across multiple institutions~\cite{Koutsoubis2024, sandhu2023medical,teo2024federated}.
A range of federated segmentation methods have been introduced to tackle key challenges such as data heterogeneity, communication inefficiency, and limited generalization across sites. Notable examples include FedEvi~\cite{Che_FedEvi_MICCAI2024}, which employs evidence-based aggregation to account for uncertainty in model updates; FedGS~\cite{schutte2024fedgs}, which utilizes gradient statistics to enhance robustness; and Fed-MENU~\cite{Xu2023FedMENU}, which integrates model ensembling with uncertainty-aware learning to improve stability. Transformer-based \cite{vaswani2017attention} methods leverage self-attention mechanisms and present high performance in distributed training for positron emission tomography (PET) and computed tomography (CT) segmentation~\cite{shiri2023multi}. Additionally,  recent approaches like FedFMS~\cite{liu2024fedfmsexploringfederatedfoundation} incorporate foundation models like SAM~\cite{kirillov2023segment} and MedSAM~\cite{MedSAM} into FL through fine-tuning, aiming to harness large-scale pretrained representations for downstream segmentation tasks in a privacy-preserving context.

Despite the overall progress, these methods present limitations in terms of practical applicability. Many rely on highly customized architectures, complex aggregation strategies, or specialized pretraining, which can pose significant barriers for researchers without profound expertise in machine learning. In contrast to nnU-Net framework, Transformer- and foundational model-based approaches offer much less flexibility in terms of computational resources, making them less feasible in typical clinical environments with constrained hardware. 

These constraints limit their out-of-the-box usability and, as a result, chances of clinical application. There remains a strong need for universal FL methods that are robust across datasets, computationally efficient, and readily applicable by a broad range of users—including those with limited resources or domain-specific requirements.

\subsection*{Heterogeneous model aggregation}\label{sec:heterogeneous_aggregation}
\hypertarget{sec:heterogeneous_aggregation}{}

Conventional FL assumes identical models and similarly distributed data across clients. However, real-world deployments often involve heterogeneous datasets and varying compute capacities. To address this, HeteroFL~\cite{Enmao21} allows training subnetworks of different widths from a shared supernetwork. While effective for standard CNNs, it is incompatible with nnU-Net, which adapts model depth, resolution, and preprocessing to dataset-specific features. FjORD~\cite{horvath2021fjord} introduces Ordered Dropout for training nested submodels, enabling dynamic scaling. Yet, it relies on a fixed architecture, limiting its applicability to nnU-Net's dataset-driven design. FedRolex~\cite{alam2022fedrolex} improves fairness by cyclically training submodels from a global network. However, it still assumes all client models are subsets of a shared template—an assumption that nnU-Net’s self-configuring framework violates. Since none of the existing methods accommodates the flexible and adaptive nature of nnU-Net, we develop a new federated approach tailored specifically to nnU-Net’s architecture and requirements.

\section*{Methods}\label{methods}
\hypertarget{sec:methods}{}

We propose two tailored methods to federate nnU-Net's pipeline for image segmentation, namely, FFE and AsymFedAvg. These two approaches are tailored towards addressing the challenges of decentralized, heterogeneous data in model training while simultaneously allowing to maintain nnU-Net's self-configuring advantages.

\subsection*{nnU-Net vs FednnU-Net}
nnU-Net is a multi-stage segmentation pipeline initiated by a preparation stage, where dataset fingerprint extraction is performed and a customized training plan is generated, which includes a tailored segmentation network architecture. This is followed by a training stage before concluding with a post-training stage that selects the best-performing models alongside optimal image post-processing techniques. To extend nnU-Net into a FL framework, we integrate our proposed methods in the respectively corresponding components of this pipeline. The first method, FFE, is incorporated into the preparation stage, ensuring that data characteristics from all participating institutions are captured and considered when generating the training plan. The second method, AsymFedAvg, operates during the training stage and facilitates the aggregation of model weight updates across nodes with heterogeneous network architectures. With the following sections providing detailed descriptions of each method, we use the terms "node" and "client" interchangeably to refer to each institution participating in the decentralized training process. We further denote the term "server" as the coordinating agent responsible for communication and aggregation of inputs from the nodes.

\subsection*{Federated Fingerprint Extraction}

The proposed FFE method enables the analysis of the federated datasets and the configuration of nnU-Net, allowing to determine a unified, one-fits-all training strategy for all data centers in the federated setup. This approach effectively approximates the configuration behavior in a centralized environment. The dataset's fingerprint is a crucial component of the nnU-Net's adaptive configuration, summarizing the key dataset characteristics that are used to determine the model configuration and preprocessing strategies. The FFE approach comprises two key steps:
\begin{itemize}
    \item \textbf{Local Fingerprint Extraction: }
    Each participating node generates their dataset fingerprint by invoking the extraction process of nnU-Net. This local fingerprint encloses the local dataset's spatial characteristics such as spacing and voxel size.
    \hfill
    
    \item \textbf{Global Fingerprint Aggregation: }
    The nodes share their local fingerprint with the server, where these get aggregated to form a global federated fingerprint \textit{z}.
    Specifically, new lists of \texttt{shapes\_after\_crop} and \texttt{spacings} are created by the concatenation of their local representations.

\end{itemize}

\begin{figure*}[t]
    \includegraphics[trim=0cm 0.5cm 0cm 1cm, clip, width=\textwidth,scale=0.35]{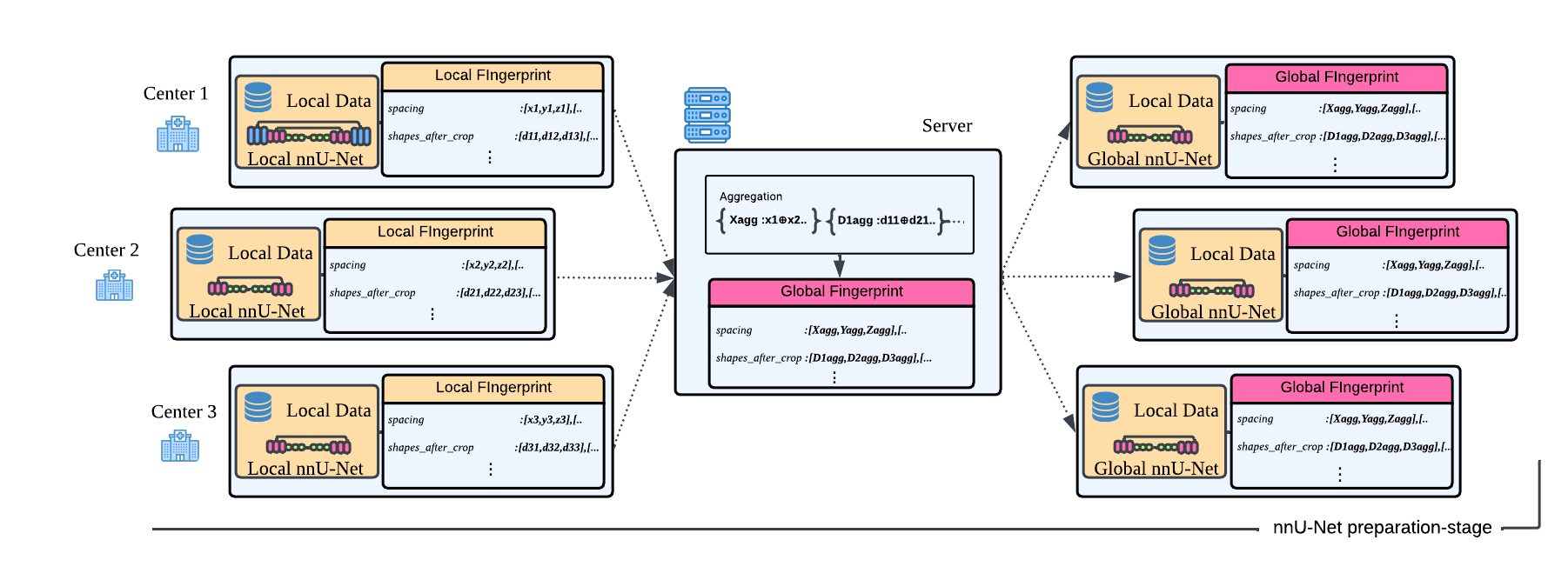}
    \caption{The Federated Fingerprint Extraction (FFE) process. Each node computes a local fingerprint from its dataset and shares it with the server, which then aggregates them into a global fingerprint. Finally, the server redistributes this global fingerprint to all nodes for local nnU-Net configuration. This process happens once at the beginning of the training stage.}
    \label{fig:FFE-overview}
\end{figure*}

Fig.~\ref{fig:FFE-overview} showcases an overview of the FFE process pipeline. As a final step, each node calculates its own nnU-Net training plan based on the combination of the received global fingerprint and the local hardware characteristics. This process ensures that all nodes will be optimizing either highly compatible or, in the case of nodes with sufficiently similar local memory and hardware constraints, identical network architectures.

\subsection*{Asymmetric Federated Averaging}

In scenarios where nnU-Net is deployed across federated nodes, variations in model architecture can arise due to differences in data characteristics or computational capabilities. To address this inherent heterogeneity, we extend the classical FedAvg approach by developing a novel Asymmetric Federated Averaging strategy (AsymFedAvg). In contrast to other heterogeneous aggregation methods mentioned in \hyperlink{sec:heterogeneous_aggregation}{Heterogeneous model aggregation}, our method allows the aggregation of models with different lengths and layer configurations. AsymFedAvg enables nodes to share only the parts of the model that are common across the rest of the federation, effectively eliminating the need to constrain all nodes to a single, identical architecture.

At each participating node, we initialize nnU-Net's adaptive configuration pipeline to generate node-specific U-Net architectures based on the aforementioned locally extracted dataset fingerprints of the centers. For illustration, a set of example configurations obtained from each data center is presented in Table~\ref{tab:nnunet_configurations_mama_mia}.

\begin{figure*}
    \centering
    \includegraphics[ width = \textwidth,scale=0.35]{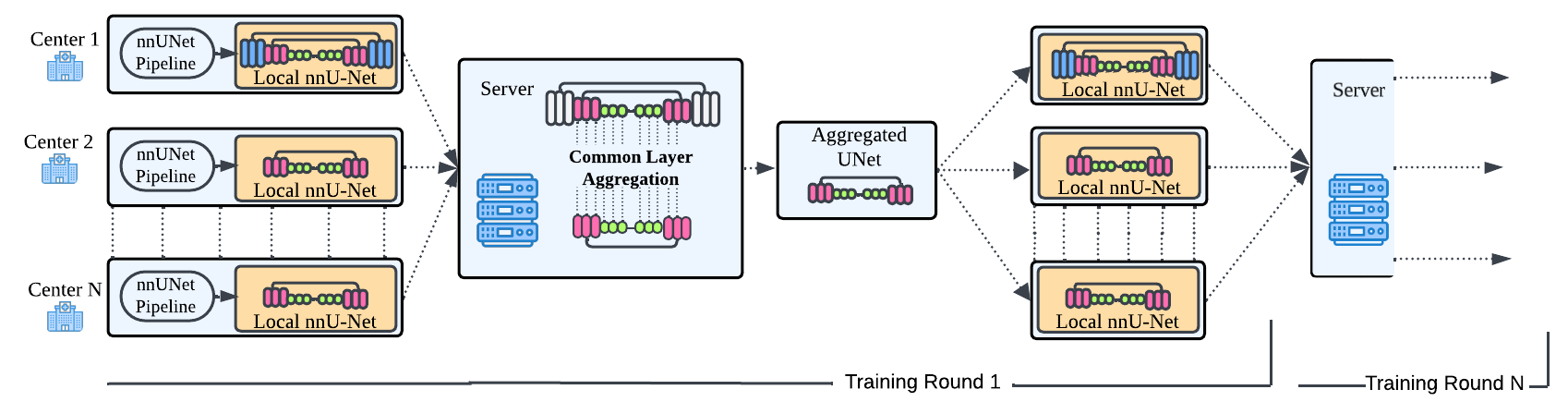}
    \caption{Overview of the Asymmetric Federated Averaging (AsymFedAvg) Training Pipeline. The nodes of the federated setup generate a local nnU-Net from the nnU-Net pipeline. During training, the server receives the local nnU-Net weights of each node and aggregates them utilizing the Aggregation Function of AsymFedAvg. At the end of each training round, the server distributes the aggregated layers of the network back into the nodes. This process is respectively repeated throughout the entire model training process.}
    \label{fig:AsymFedAvg-pipe}
\end{figure*}

\begin{table}[t]
\centering
\caption{Default nnU-Net training configurations for each center in MAMA-MIA dataset}
\label{tab:nnunet_configurations_mama_mia}
\scalebox{0.9}{%
\begin{tabular}{lcccc}
\hline
Parameter & ISPY1 & ISPY2 & DUKE & NACT \\
\hline
\multicolumn{5}{l}{2D Configuration} \\
Batch Size & 49 & 32 & 12 & 49 \\
Patch Size & [256,256] & [320,320] & [512,512] & [256,256] \\
Spacing & [0.78,0.78] & [0.68,0.68] & [0.70,0.70] & [0.70,0.70] \\
U-Net Stages & 7 & 7 & 8 & 7 \\
Features/Stage & [32-512] & [32-512] & [32-512] & [32-512] \\
\hline

\end{tabular}
}
\end{table}

\hfill

Let $\mathcal{K} = {1,\dots, K}$ be the set of participating nodes in the federation. For each node $k \in \mathcal{K}$ we define a local model state dictionary as: 
\begin{equation}
    S^k = {(l_i^k, \theta_i^k) | i \in \mathcal{L}^k},
\end{equation}

where $l_i^k$ represents the layer identifier, $\theta_i^k$ denotes the corresponding parameters for node $k$, and $\mathcal{L}^k$ is the set of layers in node $k$'s model.

We define a layer matching function $\mathcal{C}$ which identifies the common layers of the models participating in the federation as: 
\begin{equation}
    \mathcal{C}({S^k}_{k=1}^K) = {l | \forall k,j \in \mathcal{K}: l \in \mathcal{L}^k \cap \mathcal{L}^j \land \text{dim}(\theta_l^k) = \text{dim} (\theta_l^j)}.
\end{equation}

For each compatible layer $l \in \mathcal{C}({S^k}_{k=1}^K)$, the aggregated parameters $\hat{\theta}_l$ are computed as:

\begin{equation}
    \hat{\theta}_l = \frac{1}{|\mathcal{K}l|} \sum{k \in \mathcal{K}_l} \theta_l^k ,
\end{equation}

where $\mathcal{K}_l$ is the set of nodes that contain layers $l$.

The aggregated model state dictionary $\hat{S}$ is then constructed as: 
\begin{equation}  
    \hat{S} = {(l, \hat{\theta}l) | l \in \mathcal{C}({S^k}{k=1}^K)}.
\end{equation}

During each federation round $t$, the update process goes as presented in Algorithm ~\ref{alg:asymfedavg}.

\begin{algorithm*}
    \caption{Asymmetric Model Aggregation}
    \label{alg:asymfedavg}
    \begin{minipage}[t]{0.6\textwidth}
        \begin{algorithmic}[1]
            \State Each node $k$ sends its state dictionary $S^k_t$ to the server
            \State The server identifies compatible layers using the matching function $\mathcal{C}$
            \State The server computes $\hat{S}_t$ using the averaging procedure
            \For{each node $k$}
                \For{each layer $l$ in model}
                    \If{$l \in \mathcal{C}(\{S^k\}_{k=1}^K)$}
                        \State $\theta_{l,t+1}^k \gets \hat{\theta}_{l,t}$
                    \Else
                        \State $\theta_{l,t+1}^k \gets \theta_{l,t}^k$
                    \EndIf
                \EndFor
            \EndFor
        \end{algorithmic}
    \end{minipage}
    \hfill
    \begin{minipage}[t]{0.35\textwidth}
        \small
        \textbf{Notation:}
        \begin{itemize}
            \item $S^k_t$: State dictionary of node $k$ at round $t$
            \item $\mathcal{C}$: Layer compatibility function
            \item $\hat{\theta}_{l,t}$: Aggregated parameters for layer $l$
        \end{itemize}
    \end{minipage}
\end{algorithm*}

The \textbf{AsymFedAvg} allows nodes to maintain their unique architectural components while sharing knowledge via their common space of compatible layers, handling the diverse nature of medical imaging data across different centers. Fig.~\ref{fig:AsymFedAvg-pipe} depicts an overview of the AsymFedAvg pipeline and illustrates the process of updating the heterogeneous architectures in each training round.

\subsection*{Implementation}

\subsubsection*{Federated Learning Environment}\label{flower}

For the federated communication, we select the Flower framework~\cite{flower2020}, which is an open-source library designed to streamline FL tasks on a cluster of machines. Flower's high-level application programming interface (API) eliminates the need for delving into complex and time-consuming processes such as network communication between nodes and model aggregation, as it supports multiple FL optimization algorithms such as Federated Averaging (FedAvg), FedProx \cite{Li2018} and Federated SGD (FedSGD) \cite{McMahan2017}. Additionally, Flower's client-server architecture is well-suited for our multi-center setup, facilitating seamless integration of local models while maintaining centralized coordination on the server. 

\subsubsection*{FednnU-Net Framework}

Our framework maintains full functionality of the original nnU-Net while offering a federated extension that allows experimentation in both real-world and simulated environments. Focusing on usability, its modular structure enables seamless integration not only with the original nnU-Net but also with respective customized versions, allowing researchers to adapt their traditional centralized training methods to a federated setup. Key features include:

\begin{itemize}
    \item \textbf{Custom Architecture Selection}: Users can select from nnUNet's built-in architecture presets, including configurations optimized for high performance (e.g., residual encoder variants), even if they require more memory or compute.
    
    \item \textbf{Per-Client Resource Configuration}: GPU memory constraints can be individually defined for each client, enabling heterogeneous resource allocation scenarios such as in \textit{AsymFedAvg}.
    
    \item \textbf{Training Hyperparameter Control}: Users can configure learning rate schedules, loss functions, optimizer settings, and number of training epochs through the native \texttt{nnUNetTrainer} class.
    
    \item \textbf{Support for Pretrained Weights}: Initialization from pretrained weights is supported, allowing faster convergence or domain adaptation across clients.
    
    \item \textbf{Simulated and Real-World Federation}: The same codebase supports both simulated federated setups (single machine, multi-client emulation) and deployment across physically distributed clients.
\end{itemize}
This design ensures both high flexibility for research experimentation and practicality for real-world deployment across diverse clinical environments.

\subsubsection*{Distributed Setup}\label{cc}

Our experimental framework consists of 4 computational nodes with varying hardware and system specifications (Table~\ref{tab:hardware_specs}), resembling real-world clinical scenarios in low-resource countries. In particular, the availability of GPU device with 8GB of memory, is the requirement for each node in order to run the default nnU-Net training setup. The proposed framework was deployed and tested independently across all nodes, although the final model training is done on a single, multi-GPU cluster to simplify experimental control.

    \begin{table}[t]
    \centering
    \begin{minipage}{0.7\linewidth}
    \centering
    \caption{Hardware specifications of the computational nodes.}
    \label{tab:hardware_specs}
    \begin{tabular}{lccc}
    \toprule
    \textbf{Node} & \textbf{CPU (Cores/GHz)\textsuperscript{†}} & \textbf{GPU (Memory)\textsuperscript{‡}} & \textbf{OS\textsuperscript{§}} \\
    \midrule
    Node 1 & Intel i7-8700K (12/3.20)  & NVIDIA GTX 1080 (8GB)  & Ubuntu 22.04 \\
    Node 2 & Intel i7-8700K (12/3.20)  & NVIDIA GTX 1080 (8GB)  & Windows 10   \\
    Node 3 & Intel i7-9700K (8/3.60)   & NVIDIA RTX 2080S (8GB) & Ubuntu 22.04 \\
    Node 4 & Intel i7-9700  (8/3.00)   & NVIDIA RTX 2080S (8GB) & Ubuntu 22.04 \\
    \bottomrule
    \end{tabular}
    
    \vspace{1ex}
    \footnotesize
    \textsuperscript{†}CPU: Central Processing Unit. \quad
    \textsuperscript{‡}GPU: Graphics Processing Unit. \quad
    \textsuperscript{§}OS: Operating System.
    \end{minipage}
    \end{table}

\section*{Datasets}\label{data}

We evaluate our proposed methods on three distinct clinical applications (i.e., breast, cardiac, fetal), using datasets that serve as valuable testbeds due to their multi-institutional and heterogeneous cohorts. The following sections provide a detailed description of the datasets employed in each use case.

\begin{figure}
    \centering
    \includegraphics[scale=0.55]{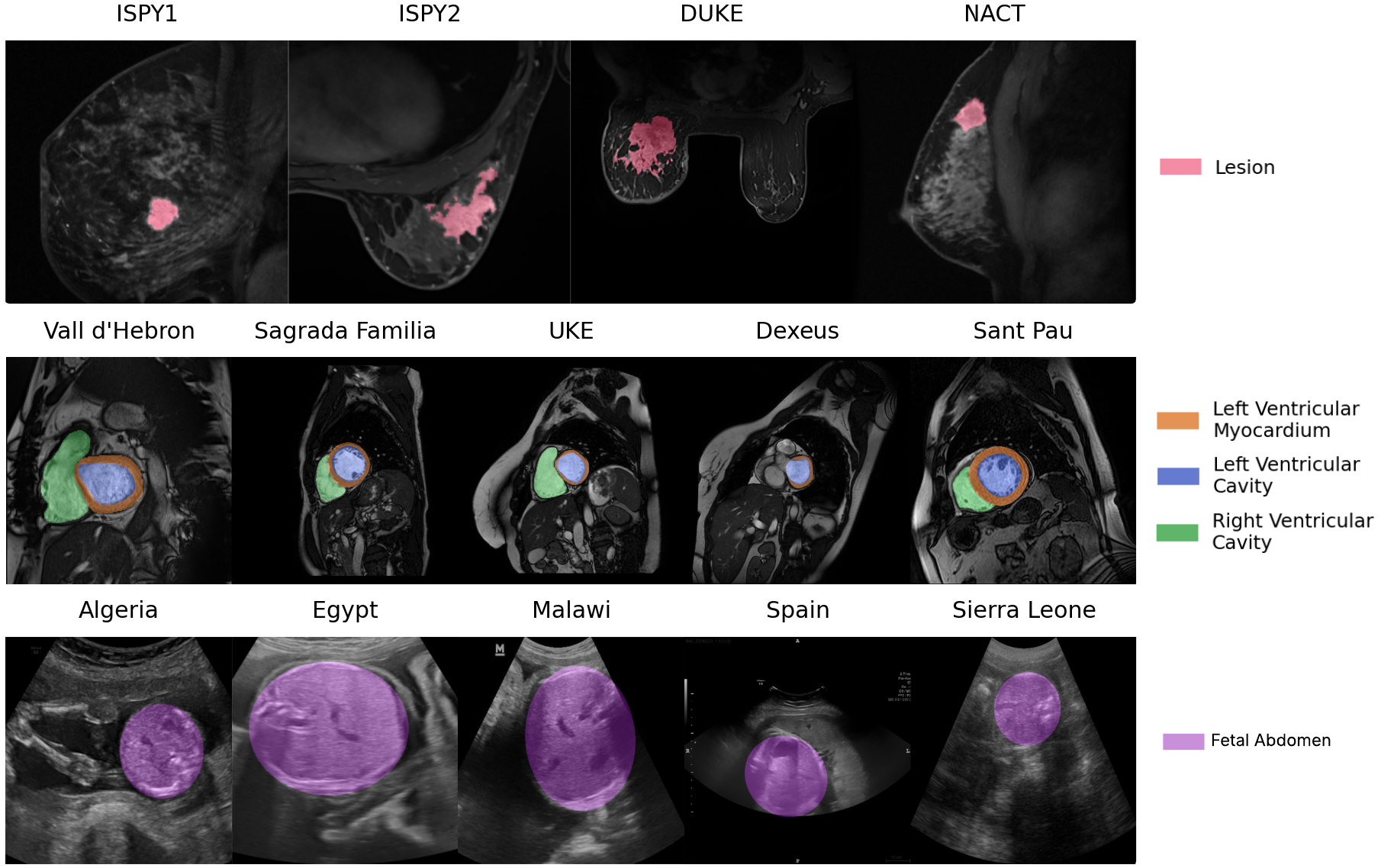}
    \caption{Example images with annotations from each data center used in this work. From the top: Breast MRI, Cardiac MRI, Fetal US}
    \label{fig:enter-label}
\end{figure}

\subsection*{Breast MRI}

\subsubsection*{MAMA-MIA}

The MAMA-MIA dataset~\cite{Garrucho2024} is the largest publicly available collection of breast dynamic contrast-enhanced MRI (DCE-MRI) cases with expert primary tumor segmentations, providing an essential resource for advancing and benchmarking artificial intelligence (AI) algorithms in breast cancer imaging. It comprises 1,506 cases acquired between 1995 and 2016 from four cohorts within The Cancer Imaging Archive (TCIA): ISPY-1 (n=171), ISPY-2 (n=980), Duke (n=291), and NACT (n=64). The annotation process focused on delineating primary tumors and non-mass-enhanced lesions within a single breast, using the first post-contrast phase as the reference—a phase typically associated with maximum enhancement of cancerous tissues that offers superior contrast between lesions and surrounding healthy tissue compared to later phases. However, significant imaging differences among these cohorts (Table~\ref{tab:mamamia-specs}) pose challenges for training robust models. For example, the NACT cohort predominantly features scans with a 2 mm slice thickness and an average of 60 slices per case, whereas Duke’s samples are mostly acquired with 1 mm thickness and include an average of 169 slices per case. Moreover, the centers used different scanner models and field strengths, further contributing to the variability in imaging protocols.

    \begin{table*}[t]
    \centering
    \caption{Summary of MRI acquisition parameters across the four centers (ISPY1, ISPY2, DUKE, and NACT) in the MAMA-MIA breast cancer dataset \cite{Garrucho2024}.}
    \begin{tabular}{lcccccc}
    \hline
    Center & Vendor & Scanner Model& \begin{tabular}[c]{@{}c@{}}Slice\\Thickness (mm)\end{tabular} & \begin{tabular}[c]{@{}c@{}}Number of\\Slices\end{tabular} & \begin{tabular}[c]{@{}c@{}}Field\\Strength (T)\end{tabular} & \begin{tabular}[c]{@{}c@{}}Number of\\Studies\end{tabular} \\
    \hline
    ISPY1 & \begin{tabular}[c]{@{}c@{}}GE (67.3\%)\\ Siemens (25.7\%)\\ Philips (7.0\%)\end{tabular} & \begin{tabular}[c]{@{}c@{}}Signa Genesis (60.2\%)\\ Other (39.8\%)\end{tabular} & 2.4 [1.5, 4.0] & 64 [44, 256] & \begin{tabular}[c]{@{}c@{}}1.5 (100.0\%)\\ 3.0 (0.0\%)\end{tabular} & 171 \\
    \hline
    ISPY2 & \begin{tabular}[c]{@{}c@{}}GE (62.3\%)\\ Siemens (25.7\%)\\ Philips (11.9\%)\end{tabular} & \begin{tabular}[c]{@{}c@{}}Signa HDxt (54.7\%)\\ Avanto (12.6\%)\\ Other (32.7\%)\end{tabular} & 2.0 [0.8, 3.0] & 106 [52, 256] & \begin{tabular}[c]{@{}c@{}}1.5 (73.0\%)\\ 3.0 (27.0\%)\end{tabular} & 980 \\
    \hline
    DUKE & \begin{tabular}[c]{@{}c@{}}GE (60.5\%)\\ Siemens (39.5\%)\end{tabular} & \begin{tabular}[c]{@{}c@{}}Other (55.7\%)\\ Avanto (24.1\%)\\ Signa HDxt (20.3\%)\end{tabular} & 1.1 [1.0, 2.5] & 169 [60, 256] & \begin{tabular}[c]{@{}c@{}}1.5 (46.7\%)\\ 3.0 (53.3\%)\end{tabular} & 291 \\
    \hline
    NACT & GE (100\%) & Signa Genesis (100\%) & 2.0 [2.0, 2.4] & 60 [46, 64] & \begin{tabular}[c]{@{}c@{}}1.5 (100.0\%)\\ 3.0 (0.0\%)\end{tabular} & 64 \\
    \hline
    \end{tabular}
    \label{tab:mamamia-specs}
    \end{table*}

\subsubsection*{EuCanImage}

    Through the European project EuCanImage~\cite{eucanimage}, we access a non-public breast MRI collection comprising 574 samples from three different countries. The EuCanImage dataset represents a multi-institutional effort to create standardized imaging biobanks for cancer research and serves here as an external test set to validate our methodology on data acquired under protocols that differ markedly from those in MAMA-MIA. Detailed specifications of the acquisition parameters are provided in Table~\ref{tab:eucanimage-specs}. Notably, while MAMA-MIA includes both axial and sagittal acquisitions, the EuCanImage collection consists exclusively of axial slices. In addition, images in EuCanImage are predominantly acquired from scanner models that differ from those represented in MAMA-MIA, and a significantly larger fraction of the scans are captured at 3T field strength. Although the spatial image characteristics are largely similar between the two datasets, these differences in scanner models, field strengths, and imaging orientations lead to distinct scan appearances, potentially challenging the robustness of segmentation algorithms.
    
    \begin{table*}[t]
    \centering
    \caption{Summary of MRI acquisition parameters across the three centers (GUMED, KAUNO, and HCB) in the EuCanImage breast MRI dataset.}
    \resizebox{\textwidth}{!}{%
    \begin{tabular}{lcccccc}
    \hline
    Center & Vendor & Scanner Model& \begin{tabular}[c]{@{}c@{}}Slice\\Thickness (mm)\end{tabular} & \begin{tabular}[c]{@{}c@{}}Pixel\\Spacing (mm)\end{tabular} & \begin{tabular}[c]{@{}c@{}}Field\\Strength (T)\end{tabular} & \begin{tabular}[c]{@{}c@{}}Number of\\Studies\end{tabular} \\
    \hline
    GUMED (Poland) & Siemens & Aera (100\%) & 1.00–1.49 & 1.00–1.49 & 1.5 (100\%) & 30 \\
    \hline
    KAUNO (Lithuania) & 
    \begin{tabular}[c]{@{}c@{}}Philips (88.4\%)\\ Siemens (11.6\%)\end{tabular} & 
    \begin{tabular}[c]{@{}c@{}}Ingenia (88.4\%)\\ Avanto (11.6\%)\end{tabular} & 
    1.50–1.99 (84.5\%) & 
    0.50–0.99 (96.6\%) & 
    \begin{tabular}[c]{@{}c@{}}1.5 (11.6\%)\\ 3.0 (88.4\%)\end{tabular} & 
    232 \\
    \hline
    HCB (Spain) & \begin{tabular}[c]{@{}c@{}}GE (69.2\%)\\ Siemens (30.8\%)\end{tabular} & \begin{tabular}[c]{@{}c@{}}Signa HDxt (63.8\%)\\ Aera (22.8\%)\\ Others (13.4\%)\end{tabular} & 2.00–2.49 (87.2\%) & 0.50–0.99 (98.1\%) & \begin{tabular}[c]{@{}c@{}}1.5 (90.7\%)\\ 3.0 (9.3\%)\end{tabular} & 312 \\
    \hline
    \end{tabular}
    }
    \label{tab:eucanimage-specs}
    \end{table*}
    
\subsection*{Cardiac MRI}

\subsubsection*{M\&Ms} 

To further validate our approach, we leverage the Multi-center, Multi-vendor \& Multi-disease (M\&Ms) cardiac MRI datasets. As a part of our study, we curate a merged version of datasets that were used in the M\&Ms challenges \cite{Campello2021,Isla2023} by excluding the overlapping samples based on anonymized patient IDs provided by each institution. This curation allows to reach the full potential of this data collection without the risk of replicated samples. The final, accumulated set consists of 543 unique, short-axis cardiac MRI scans with expert annotations from 5 distinct international institutions. We make this collection publicly available.

The M\&Ms datasets are particularly valuable for validation due to their inherent heterogeneity (Table~\ref{tab:mnms_specs}), they include images from four different vendors (Siemens, GE, Philips, Canon), various scanner models (from older systems like Symphony to newer ones like Vantage Orian), and different field strengths (1.5T and 3.0T). This multi-center, multi-vendor nature mirrors the real-world variability in clinical settings, making these datasets ideal for testing the generalizability of our approach.

The technical heterogeneity across centers is evident in variations of scanner models, field strengths and different acquisition protocols. This diversity presents similar challenges to those encountered in our MAMA-MIA analysis, particularly in terms of standardization and harmonization of imaging parameters.

\begin{table*}[t]
\centering
\caption{Average specifications of images acquired in different centers across the merged collection of M\&Ms 1 \cite{Campello2021} and M\&Ms 2 \cite{Isla2023} challenge datasets. }
\resizebox{\textwidth}{!}{%

\begin{tabular}{lcccccc}
\hline
Center & Vendor & Scanner Model& \begin{tabular}[c]{@{}c@{}}Slice\\Thickness (mm)\end{tabular} & \begin{tabular}[c]{@{}c@{}}Number of\\Slices\end{tabular}  &\begin{tabular}[c]{@{}c@{}}Field\\Strength (T)\end{tabular} & \begin{tabular}[c]{@{}c@{}}Number of\\Studies\end{tabular} \\

\hline
Hospital Vall d'Hebron & Siemens & \makecell[c]{Symphony TIM (63.18\%) \\ Avanto Fit (15.48\%) \\ Other (21.34\%)} & 9.5 & 13  & \makecell[c]{1.5 (97.91\%) \\ 3.0 (2.09\%)} & 239 \\
\hline
\makecell[l]{Clinica Sagrada \\ Familia} & Philips & Achieva & 10 & 13  & 1.5 & 98 \\
\hline
\makecell[l]{Universitätsklinikum \\ Hamburg-Eppendorf} & Philips & Achieva & 10 & 11  & 1.5 & 51 \\
\hline
Hospital Universitari Dexeus & GE & \makecell[c]{Signa Excite (75.24\%) \\ Signa HDxt (23.81\%) \\ Signa Explorer (0.95\%)} & 9.9 & 12  & \makecell[c]{1.5 (88.57\%) \\ 3.0 (11.43\%)} & 105 \\
\hline
Hospital Sant Pau & Canon & Vantage Orian & 10 & 13  &1.5 & 50 \\
\hline

\end{tabular}
}
\label{tab:mnms_specs}
\end{table*}

\subsubsection*{ACDC} The Automated Cardiac Diagnosis Challenge (ACDC) dataset consists of 150 cardiac short-axis MRI examinations acquired at the University Hospital of Dijon (France) \cite{Bernard2018}. The dataset is evenly distributed across five groups (30 cases each): normal subjects (NOR), patients with myocardial infarction (MINF), dilated cardiomyopathy (DCM), hypertrophic cardiomyopathy (HCM), and abnormal right ventricle (RV). This dataset is particularly valuable for validation due to its well-defined pathological diversity and standardized acquisition protocols. It is used for external validation of models trained on M\&Ms dataset.

\subsection*{Fetal Ultrasound}
\subsubsection*{AIMIX}

With a focus on resource-constrained environments, the \textbf{AIMIX}~\cite{aimix} dataset was developed to advance inclusive image analysis in prenatal care. This dataset addresses critical challenges, including the scarcity of African medical data, limited computational resources, and a lack of clinical expertise. The dataset comprises pregnancy blind-sweeps 2D ultrasound scans, along with associated patient metadata such as age and ethnicity. The data span multiple international sites, including \textbf{Algeria, Egypt, Malawi and Spain}, reflecting diverse geographic and healthcare settings. 

In this study, we focus on abdominal segmentation, a crucial step in prenatal ultrasound analysis. To enable the evaluation on this multi-center dataset, we provide a set of 308 standard-plane images with masks corresponding to the fetal abdomen. Our data processing workflow involved detecting the optimal abdominal plane from each blind-sweep and segmenting it using a 4-point ellipse-fitting algorithm \cite{Carneiro2008}, utilizing a semi-automatic software by a researcher with more than five years of experience and reviewed by an expert clinician. Table~\ref{tab:fetal-dataset-specs} lists the number of samples and the mean resolution for each center.

\begin{table}[t]
\centering
\caption{Summary of the AIMIX and ACOUSLIC-AI datasets, including number of samples, mean resolution, nnU-Net patch size, and UNet stages for each dataset.}
\begin{tabular}{lcccc}
\hline
Center &  \begin{tabular}[c]{@{}c@{}}Mean Resolution\\ (h x w)\end{tabular} & \begin{tabular}[c]{@{}c@{}}Number of \\ Studies \end{tabular} \\
\hline
Algeria & 418x500 & 25 \\
Egypt & 450x400 & 20 \\
Malawi & 400x500 & 25 \\
Spain & 1228x1276 & 60 \\
SierraLeone (External) & 561x743 & 178 \\
\hline
\end{tabular}
\label{tab:fetal-dataset-specs}
\end{table}

\subsubsection*{ACOUSLIC-AI Sierra Leone} 

For the external validation, we utilize the publicly available samples from \textbf{ACOUSLIC-AI} Challenge dataset~\cite{sappia_2024_12697994}. We use a collection of 178 cases collected in Sierra Leone from three Public Health Units (PHUs). 
With a resolution of 744×562 pixels, the images from Sierra Leone sit between the low-end devices used in Algeria and Malawi and the high-end scanners employed in Spain. This intermediate imaging quality makes the dataset an ideal candidate for assessing model generalizability across varying image qualities and clinical environments, ensuring robust validation of AI performance.

\section*{Experiments}\label{experiments}

\subsection*{Experimental setups}

We adopt the vanilla nnU-Net architecture in both local and centralized training setups as the primary baselines for evaluating the performance of our proposed FednnU-Net framework. This choice is grounded in the fact that nnU-Net has established itself as a strong, standardized, and adaptive baseline for medical image segmentation tasks across a wide range of datasets~\cite{nnunet_vs_other_unets}. By using the same framework and memory constraints in local (per-institution) and centralized (pooled data) settings, we ensure a fair and controlled comparison that isolates the effect of the federation process itself, without introducing variability due to methodological, architectural or computational resource differences.

This experimental setup allows for a clear evaluation of how well the federated setting bridges the gap between privacy preservation and segmentation accuracy, thus aligning with the core motivation of our study. To this end, the training process is conducted and empirically evaluated across four different setups: 
\begin{itemize}
    \item \textbf{Local Training:}~
    Each node independently trains its own nnU-Net model using only its local dataset. This setup serves as a baseline, reflecting the performance of a node/institution without participation in the federation. 
    
    \item \textbf{Centralized Training: }~
    All data is aggregated into a single dataset, which is used to train a centralized nnU-Net model. This setup represents the theoretical upper bound of performance, though it is typically infeasible 
    and often not recommended in real-world scenarios due to significant privacy risks and data-sharing constraints.
    
    \item \textbf{Federated Fingerprint Training: }~
     Utilizing our FFE approach (see \hyperlink{sec:methods}{Methods}), the local dataset fingerprints from all nodes are aggregated at the central node into one federated fingerprint and distributed back to each node. As a result, each training site obtains an identical network architecture during the nnU-Net's self-configuration process and enables the utilization of a standard FedAvg aggregation method. Moreover, this approach approximates nnU-Net's behavior in the centralized training setup.

    \item \textbf{Asymmetric Aggregation Training: }~
    Implemented using our \textbf{AsymFedAvg} strategy, this allows nodes to maintain their original nnU-Net configured architectures while sharing compatible parts of their networks. 
    
    \end{itemize}

For our experimentation, we maintain nnU-Net's default configuration parameters as described in \cite{Isensee2021}. We test the 2D U-Net architecture on breast MRI and fetal ultrasound data and the 3D architecture on a cardiac MRI use case, consistently employing the default training length of 1000 epochs and using the last model checkpoint for evaluation across all federated nodes. In federated scenarios, all models are aggregated after each epoch and sent back to the clients.
Following other recent works on medical image segmentation \cite{Isensee2021, UNETR, nnformer}, we evaluate the models with Dice Similarity Coefficient (DSC) and 95\% Hausdorff distance (HD95). 
For internal validation, we report results as the average of a 5-fold cross-validation, while for external testing, we follow the original nnU-Net approach by generating ensemble predictions from all five folds. Moreover, each experiment uses identical data splits—for example, the n-th fold in the centralized setup corresponds directly to the set of n-th folds from centers used in the local setup.
To assess statistical significance between segmentation methods, we apply the two-tailed Wilcoxon signed-rank test on paired DSC scores with 5\% significance threshold.
By adhering to these protocols, we try to minimize the influence of other factors on the final results, ensuring that any observed performance variations are primarily due to our federated learning adaptations rather than modifications to the base framework.

\subsection*{Results}\label{results}

\begin{table*}[!t]
\centering
\caption{Quantitative results for multi-center MAMA-MIA and EuCanImage datasets. HD95 values reported in \textit{mm}. Best results for privacy-preserving methods are in bold.}
\small
\setlength{\tabcolsep}{2pt}
\scalebox{0.94}{%
\begin{tabular}{lccccccccccccccc}
\hline
Method & \multicolumn{2}{c}{ISPY1} & \multicolumn{2}{c}{ISPY2} & \multicolumn{2}{c}{DUKE} & \multicolumn{2}{c|}{NACT} & \multicolumn{2}{c}{\makecell{external \\ GUMED}} & \multicolumn{2}{c}{\makecell{external \\ KAUNO}} & \multicolumn{2}{c}{\makecell{external \\ HCB}} \\
\cmidrule(lr){2-3}  \cmidrule(lr){4-5}  \cmidrule(lr){6-7}  \cmidrule(lr){8-9} \cmidrule(lr){10-11} \cmidrule(lr){12-13} \cmidrule(lr){14-15}
Metrics & DSC $\uparrow$ & HD95 $\downarrow$ & DSC $\uparrow$ & HD95 $\downarrow$ & DSC $\uparrow$ & HD95 $\downarrow$ & DSC $\uparrow$ & HD95 $\downarrow$ & DSC $\uparrow$ & HD95 $\downarrow$ & DSC $\uparrow$ & HD95 $\downarrow$ & DSC $\uparrow$ & HD95 $\downarrow$ \\
\hline
nnU-Net (local) & 0.612 & 17.60 & 0.739 & 19.03 & 0.616 & 33.56 & 0.551 & 23.24 & 0.579 & 35.64 & 0.636 & \textbf{31.72} & 0.544 & 46.30 \\
FednnU-Net AsymFedAvg  & 0.634 & 16.69 & 0.739 & 18.15 & 0.621 & \textbf{28.70} & 0.620 & 21.79 & 0.557 & 51.43 & 0.637 & 32.43 & 0.551 & 49.10 \\
FednnU-Net FFE & \textbf{0.644} & \textbf{16.20} & \textbf{0.742}  & \textbf{17.48} & \textbf{0.653} & 39.54 & \textbf{0.660} & \textbf{17.49} & \textbf{0.582} & \textbf{28.83} & \textbf{0.645} & 33.52 & \textbf{0.565} & \textbf{40.41} \\
\hline
nnU-Net (centralized) & 0.654 & 17.28 & 0.750 & 19.84 & 0.657 & 58.99 & 0.675 & 17.35 & 0.582 & 52.44 & 0.644 & 41.17 & 0.569 & 57.98 \\
\hline

\end{tabular}
}
\label{tab:mama_mia_results}
\end{table*}

\begin{table*}[!t]
\centering
\caption{Quantitative results for multi-center M\&Ms 
 and ACDC datasets. HD95 values reported in \textit{mm}. Best results for privacy-preserving methods are in bold.}
\small
\setlength{\tabcolsep}{3pt}
\scalebox{0.94}{%
\begin{tabular}{lcccccccccccc}
\hline
Method & \multicolumn{2}{c}{\makecell{Hospital \\ Vall d’Hebron}} & \multicolumn{2}{c}{\makecell{Clinica \\ Sagrada Familia}} & \multicolumn{2}{c}{\makecell{Universitätsklinikum \\ Hamburg-Eppendorf}} & \multicolumn{2}{c}{\makecell{Hospital \\ Dexeus}} & \multicolumn{2}{c|}{\makecell{Hospital \\ Sant Pau}} & \multicolumn{2}{c}{\makecell{external \\ ACDC}} \\
\cmidrule(lr){2-3}  \cmidrule(lr){4-5}  \cmidrule(lr){6-7}  \cmidrule(lr){8-9} \cmidrule(lr){10-11} \cmidrule(lr){12-13}
Metrics & DSC $\uparrow$ & HD95 $\downarrow$ & DSC $\uparrow$ & HD95 $\downarrow$ & DSC $\uparrow$ & HD95 $\downarrow$ & DSC $\uparrow$ & HD95 $\downarrow$ & DSC $\uparrow$ & \multicolumn{1}{c|}{HD95 $\downarrow$} & DSC $\uparrow$ & HD95 $\downarrow$ \\
\hline
nnU-Net (local) & 0.908 & 4.27 & 0.912 & \textbf{3.62} & 0.902 & 6.56 & 0.896 & 4.31 & 0.900 & \multicolumn{1}{c|}{4.34} & 0.887 & 4.79\\
FednnU-Net AsymFedAvg  & 0.909 & 4.20 & 0.913 & 3.70 & 0.907 & 4.23 & 0.897 & 4.29 & 0.902 & \multicolumn{1}{c|}{4.07} & 0.889 & 4.58\\
FednnU-Net FFE & \textbf{0.911} & \textbf{4.05} & \textbf{0.914} & 3.67 & \textbf{0.922} & \textbf{3.30} & \textbf{0.904} & \textbf{3.98} & \textbf{0.904} & \multicolumn{1}{c|}{\textbf{3.87}} & \textbf{0.897} & \textbf{3.87}\\
\hline
nnU-Net (centralized) & 0.910 & 4.20 & 0.914 & 3.53 & 0.923 & 3.26 & 0.905 & 3.87 & 0.907 & \multicolumn{1}{c|}{3.81} & 0.898 & 3.89\\
\hline
\end{tabular}
}
\label{tab:mnms_results}
\end{table*}

\begin{table*}[!t]
\centering
\caption{Quantitative results for multi-center AIMIX and ACOUSLIC-AI datasets. HD95 values reported in \textit{pixels} due to missing spacing information. Best results for privacy-preserving methods are in bold.}
\setlength{\tabcolsep}{4pt}
\scalebox{0.85}{%
\begin{tabular}{lcccccccccc}
\hline
Method & \multicolumn{2}{c}{Algeria} & \multicolumn{2}{c}{Egypt} & \multicolumn{2}{c}{Malawi} & \multicolumn{2}{c|}{Spain} & \multicolumn{2}{c}{\makecell{external \\ Sierra Leone}} \\
\cmidrule(lr){2-3}  \cmidrule(lr){4-5}  \cmidrule(lr){6-7}  \cmidrule(lr){8-9} \cmidrule(lr){10-11} 
Metrics & DSC $\uparrow$ & HD95 $\downarrow$ & DSC $\uparrow$ & HD95 $\downarrow$ & DSC $\uparrow$ & HD95 $\downarrow$ & DSC $\uparrow$ & \multicolumn{1}{c|}{HD95 $\downarrow$} & DSC $\uparrow$ & HD95 $\downarrow$ \\
\hline
nnU-Net (local) & 0.952& 11.87 & 0.931& 24.21 & 0.926& 26.83 & 0.942& \multicolumn{1}{c|}{29.82} & 0.831 & 53.20\\
FednnU-Net AsymFedAvg  & 0.951& 11.16 & 0.941& 18.29 & 0.939& 16.20 & \textbf{0.944}& \multicolumn{1}{c|}{\textbf{ 25.42}} & \textbf{0.851} & \textbf{43.44}\\
FednnU-Net FFE & \textbf{0.955}& \textbf{10.23} & \textbf{0.954}& \textbf{14.56} & \textbf{0.943}& \textbf{15.64} & 0.938 & \multicolumn{1}{c|}{47.58} & 0.809 & 46.57\\
\hline
nnU-Net (centralized) & 0.954 & 10.23 & 0.953& 14.50 & 0.943& 15.55 & 0.932& \multicolumn{1}{c|}{55.76} & 0.804 & 54.34\\
\hline

\end{tabular}}
\label{tab:fetal_results}
\end{table*}

\subsubsection*{Breast MRI}

Table~\ref{tab:mama_mia_results} summarizes segmentation performance on the MAMA-MIA dataset. Both FFE and AsymFedAvg outperform local training across all centers in DSC and HD95. FFE achieves average DSC improvements over local models on ISPY1 (3.2\%, p=0.009), ISPY2 (0.3\%, p=0.858), DUKE (3.7\%, p=0.006), and NACT (10.9\%, p<0.001), also closely matching centralized results in DSC and surpassing them in HD95. Notably, AsymFedAvg yields the best HD95 on DUKE, reflecting its advantage in handling heterogeneous data (DUKE’s pixel size is 2× smaller; see Table~\ref{tab:mamamia-specs}). Fig.~\ref{fig:cross-center-overview} highlights this heterogeneity, with ISPY2 models generalizing better to DUKE due to similar resolutions. Sample predictions (Fig.~\ref{fig:qualitative_results_breast}) show marked improvements for NACT and DUKE using federated methods. External validation on EuCanImage confirms FFE's generalization, achieving DSC comparable to centralized training. KAUNO center results are consistent with MAMA-MIA, while GUMED’s drop likely reflects annotation discrepancies which are illustrated in Fig.~\ref{fig:qualitative_results}.

\subsubsection*{Cardiac MRI}

For M\&Ms, federated methods outperform local models on both metrics, except for Clinica Sagrada Familia, where HD95 is slightly better locally. FFE consistently achieves the best DSC scores across internal and external datasets (combined significance using Fischer's method: p<0.001), while AsymFedAvg shows limited benefit—likely due to the dataset's homogeneity, where unified architectures are more suitable. Cross-center results (Fig.~\ref{fig:cross-center-overview}) support this, with models generalizing well across unseen centers. Importantly, FFE matches centralized performance despite additional privacy constraints. Sample predictions (Fig.~\ref{fig:qualitative_results_cardiac}) show consistent results for central slices, with more variability at the apex (ACDC) and atria (Vall d'Hebron).

\subsubsection*{Fetal Ultrasound}

\begin{figure*}
    \centering
    \includegraphics[scale=0.23]{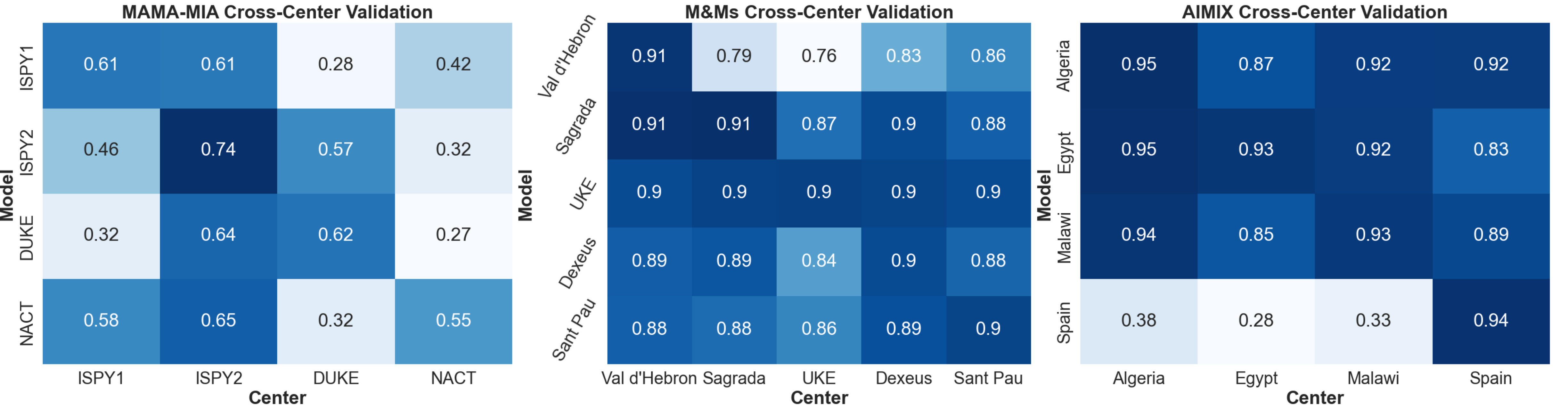}
    \caption{Cross-center DSC scores for isolated model training predictions across multiple centers.}
    \label{fig:cross-center-overview}
    \label{fig:mama-cross-center}
    \label{fig:mnm-cross-center}
    \label{fig:aimix-cross-center}
\end{figure*}

\begin{figure*}
    \centering
    \includegraphics[scale=0.23]{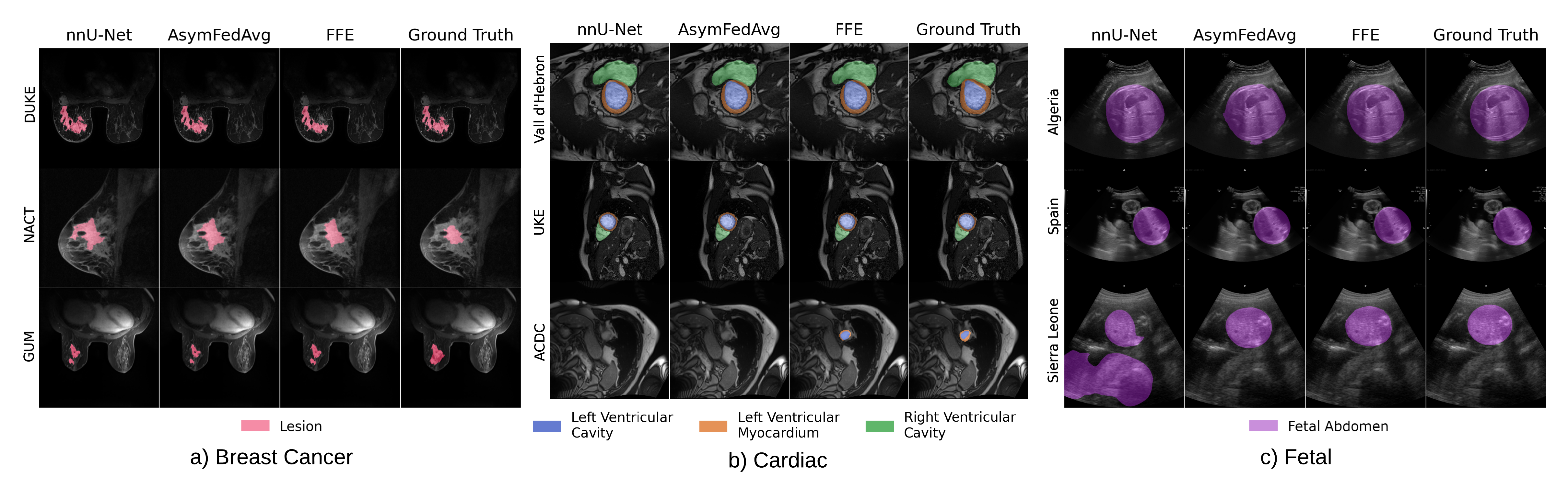}
    \caption{Sample predictions of each method for datasets used in the study.}
    \label{fig:qualitative_results}
    \label{fig:qualitative_results_breast}
    \label{fig:qualitative_results_cardiac}
    \label{fig:qualitative_results_fetal}
\end{figure*}

In the fetal abdomen segmentation task, FFE and AsymFedAvg differ markedly (Table~\ref{tab:fetal_results}). Strong heterogeneity (capture devices, resolution; Table~\ref{tab:fetal-dataset-specs}) impairs generalization from Algeria, Egypt, and Malawi to higher-resolution Spanish samples (Fig.~\ref{fig:aimix-cross-center}). Unified architectures (FFE) are less optimal here: allowing Spain to retain its native architecture (AsymFedAvg) improves performance for external centers like Sierra Leone, notably the DSC raised to 0.851 (p < 0.001). Fig.~\ref{fig:qualitative_results_fetal} shows that local models often misidentify abdomen regions, especially for Sierra Leone, whereas federated approaches improve delineation.

\section*{Discussion} \label{discussion}
\subsection*{Key Contributions and Performance}
This study introduces FednnU-Net, a federated learning framework for 2D and 3D medical image segmentation that combines privacy preservation with high performance through two key innovations: Federated Fingerprint Extraction for standardized architecture sharing, and AsymFedAvg for flexible aggregation across heterogeneous node-specific models.
Our experiments across three diverse medical imaging clinical tasks in (i) breast MRI (MAMA-MIA
), (ii) cardiac MRI (M\&Ms)
, and (iii) fetal ultrasound (AIMIX) - show that FednnU-Net reliably matches or surpasses the performance of 
centralized strategies across settings and modalities. These results add to a growing body of evidence illustrating that distributed model optimization can achieve comparable accuracy while simultaneously fulfilling the requirements of complex real-world clinical collaborations, including those related to data privacy and deviations in institution-specific protocols. 
\subsection*{Towards Trustworthy AI in Healthcare}
Moreover, our findings underscore the critical need for, and the potential of, methods that effectively handle data heterogeneity during federated model training. To this end, our framework supports site-specific model adaptation and evaluation in line with the FUTURE-AI guidelines for trustworthy AI in healthcare \cite{lekadir2025future}, particularly the \nth{4} Universality Recommendation, which calls for assessing local clinical validity amid variations in populations, equipment, workflows, and end users. FednnU-Net addresses this by enabling center-specific model customization, allowing institutions to tailor models to their local data characteristics. This approach ensures alignment with local workflows while maintaining strong performance across diverse populations, thereby promoting trust and robustness in decentralized clinical environments.

While some federated segmentation methods report high accuracy by tailoring solutions to narrow, task-specific settings, this work focuses on delivering an accessible, privacy-preserving framework that emphasizes ease of use and adaptability. By prioritizing broad applicability and real-world viability, the framework is positioned to support diverse research settings and facilitate widespread adoption. 

Indeed, the ongoing shift towards privacy-preserving, decentralized deep learning solutions in healthcare is critical for protecting patient confidentiality \cite{teo2024federated}. In addition to this, the FUTURE-AI General 2 guideline states the necessity of implementing rigorous data protection throughout the AI lifecycle, including the use of privacy-enhancing technologies and secure data governance. In the context of FL with FednnU-Net, data protection is operationalized by ensuring that patient data never leaves its local institutions of provenance, reducing the risk of sensitive information leakage during collaborative model training. To this end, FednnU-Net addresses critical privacy and security concerns inherent in decentralized healthcare AI solutions, fostering trust among stakeholders and ensuring compliance with ethical and regulatory standards.
While FednnU-Net addresses critical requirements for privacy and local adaptation in line with FUTURE-AI guidelines, several important aspects still warrant further investigation:

\begin{enumerate}
    \item \textbf{Scalability and Client Diversity:}  Future work is to assess FednnU-Net across an even larger, geographically dispersed federation with diverse patient populations \cite{lekadir2025future} to study convergence dynamics at international, multi-institutional scales.
    \item \textbf{Heterogeneous Computational Resources:} While our framework already supports diverse node capabilities, further systematic ablation studies under varied hardware conditions, including low-resource settings \cite{fabila2025federated}, are motivated towards showing the versatility for clinical adoption of the framework.
    \item \textbf{Enhanced Privacy Mechanisms:} FL improves data privacy by design, but it still does not guarantee full security. Attacks such as model inversion~\cite{geiping2020inverting} or membership inference~\cite{nasr2019comprehensive} can still extract sensitive information from shared model updates. Integrating differential privacy~\cite{dp_n_federated_learning,kaissis2020secure,ziller2021differentially,riess2024complex}, synthetic data generation \cite{raggio2025fedsynthct, osuala2023data,osuala2024enhancing}, and secure multiparty computation\cite{zhou2024secure} could further reduce the risk of data leakage, even in aggregated updates.
\end{enumerate}

\subsection*{Conclusion}
FednnU-Net fills a critical gap by balancing segmentation performance, ease of use, and privacy preservation. As FL gains adoption in healthcare \cite{teo2024federated}, our FednnU-Net framework is a key component in enabling the translation of high-performance deep learning into privacy-compliant, scalable clinical solutions.
Our framework enables researchers and practitioners to easily integrate nnU-Net, along with other dedicated frameworks and innovations built upon it, such as for instance Bayesian uncertainty estimation \cite{vorberg2024bayesian}, continual learning \cite{gonzalez2023lifelong}, and curriculum learning extensions \cite{fischer2024progressive}, into FL setups, thereby broadening their applicability under privacy constraints and supporting the increasingly prevalent multi-center collaborations. Continued evaluation, targeted enhancements, and the combination with such innovations is to further strengthen the versatility and robustness of the framework, paving the way for widespread, trustworthy deployment across diverse clinical environments.

\subsection*{Data availability}
Our framework's code is publicly available at https://github.com/faildeny/FednnUNet. The MAMA-MIA dataset access information is available at https://github.com/LidiaGarrucho/MAMA-MIA. The access to the updated version of M\&Ms1 The M\&Ms2 datasets and annotations for fetal US will be provided through https://github.com/faildeny/FednnUNet upon the publication of this article. For any queries regarding the data used in this study, please contact the corresponding author.

\section*{Funding information}
This work received funding from the European Union’s Horizon Europe research and innovation programme under Grant Agreement No. 101057849 (DataTools4Heart). This work has been supported by the European Union’s research and innovation programmes: Horizon Europe under Grant Agreement No. 101057699 (RadioVal) and Grant Agreement No. 101044779 (AIMIX), Horizon 2020 under Grant Agreement No. 952103 (EuCanImage).

\section*{Author contributions statement}
The study was conceived by G.S. and designed by G.S., C.M.-I., and K.L. G.S. developed the methods, wrote the program code, and conducted the experimental work. F.A. and C.M.-I. contributed to code development and experimental work. F.A., D.A.K., R.O., and E.R.P. contributed to manuscript writing. G.S. led the writing of the manuscript. G.S., F.A., D.A.K., and R.O. revised the article. C.M.-I., F.A., L.G., D.A.K., and E.R.P. worked on data preparation. M.B., K.G., X.B., and P.J. provided the imaging data and annotations for this study. M.B., K.G., X.B., P.J., C.M.-I., E.R.P., O.D., K.K., and K.L. reviewed the manuscript. K.K. and K.L. supervised the study. K.L. provided funding for the study. All authors have read and approved the final manuscript.

\section*{Additional information}
\textbf{Competing interests} The authors declare no competing interests.

\bibliographystyle{ieeetr}
\bibliography{main}

\begin{thebibliography}{10}

\bibitem{azad2024medical}
A.~Reza {\em et~al.}, ``Medical image segmentation review: The success of u-net,'' {\em IEEE Transactions on Pattern Analysis and Machine Intelligence}, 2024.

\bibitem{xu_2022_a}
X.~Yang {\em et~al.}, ``A medical image segmentation method based on multi-dimensional statistical features,'' {\em Frontiers in Neuroscience}, vol.~16, 09 2022.

\bibitem{Bernard2018}
O.~Bernard {\em et~al.}, ``Deep learning techniques for automatic mri cardiac multi-structures segmentation and diagnosis: Is the problem solved?,'' {\em IEEE Transactions on Medical Imaging}, vol.~37, no.~11, pp.~2514--2525, 2018.

\bibitem{hollon2020near}
T.~C. Hollon {\em et~al.}, ``Near real-time intraoperative brain tumor diagnosis using stimulated raman histology and deep neural networks,'' {\em Nature medicine}, vol.~26, pp.~52--58, 2020.

\bibitem{esmaeili2018direction}
E.~Morteza, S.~A. Line, B.~E. Magnus, S.~Ole, and R.~Ingerid, ``The direction of tumour growth in glioblastoma patients,'' {\em Scientific reports}, vol.~8, no.~1, p.~1199, 2018.

\bibitem{LeCun2015}
Y.~LeCun, Y.~Bengio, and G.~Hinton, ``Deep learning,'' {\em Nature}, vol.~521, pp.~436--444, 5 2015.

\bibitem{Fukushima1980}
K.~Fukushima, ``Neocognitron: A self-organizing neural network model for a mechanism of pattern recognition unaffected by shift in position,'' {\em Biological Cybernetics}, vol.~36, pp.~193--202, 4 1980.

\bibitem{Isensee2021}
F.~Isensee, P.~F. Jaeger, S.~A.~A. Kohl, J.~Petersen, and K.~H. Maier-Hein, ``nnu-net: a self-configuring method for deep learning-based biomedical image segmentation,'' {\em Nature Methods}, vol.~18, pp.~203--211, 2 2021.

\bibitem{nnUNet_as_basis}
J.~Ma, ``Cutting-edge 3d medical image segmentation methods in 2020: Are happy families all alike?.'' arXiv: 2101.00232, 2021.

\bibitem{Ise_nnUNet_MICCAI2024}
F.~Isensee {\em et~al.}, ``{ nnU-Net Revisited: A Call for Rigorous Validation in 3D Medical Image Segmentation },'' in {\em proceedings of Medical Image Computing and Computer Assisted Intervention -- MICCAI 2024}, vol.~LNCS 15009, Springer Nature Switzerland, October 2024.

\bibitem{ronneberger2015u}
O.~Ronneberger, P.~Fischer, and T.~Brox, ``U-net: Convolutional networks for biomedical image segmentation,'' in {\em Medical image computing and computer-assisted intervention–MICCAI 2015: 18th international conference, Munich, Germany, October 5-9, 2015, proceedings, part III 18}, pp.~234--241, 2015.

\bibitem{Liang2024}
B.~Liang {\em et~al.}, ``Automatic segmentation of 15 critical anatomical labels and measurements of cardiac axis and cardiothoracic ratio in fetal four chambers using nnu-netv2,'' {\em BMC Med. Inform. Decis. Mak.}, vol.~24, 05 2024.

\bibitem{choi2024deep}
Y.~Choi, J.~Bang, S.-Y. Kim, M.~Seo, and J.~Jang, ``Deep learning--based multimodal segmentation of oropharyngeal squamous cell carcinoma on ct and mri using self-configuring nnu-net,'' {\em European Radiology}, vol.~34, no.~8, pp.~5389--5400, 2024.

\bibitem{nnunet_vs_other_unets}
L.~Huang, A.~Miron, K.~Hone, and Y.~Li, ``{ Segmenting Medical Images: From UNet to Res-UNet and nnUNet },'' in {\em 2024 IEEE 37th International Symposium on Computer-Based Medical Systems (CBMS)}, (Los Alamitos, CA, USA), pp.~483--489, IEEE Computer Society, June 2024.

\bibitem{McMahan2017}
H.~B. McMahan, E.~Moore, D.~Ramage, S.~Hampson, and B.~A. y~Arcas, ``Communication-efficient learning of deep networks from decentralized data,'' in {\em International Conference on Artificial Intelligence and Statistics}, 2016.

\bibitem{fischer2024progressive}
S.~M. Fischer, L.~Felsner, R.~Osuala, J.~Kiechle, D.~M. Lang, J.~C. Peeken, and J.~A. Schnabel, ``Progressive growing of patch size: Resource-efficient curriculum learning for dense prediction tasks,'' in {\em International Conference on Medical Image Computing and Computer-Assisted Intervention}, pp.~510--520, Springer, 2024.

\bibitem{wang2024towards}
C.~Wang, Y.~Liu, F.~Wang, C.~Zhang, Y.~Wang, M.~Yuan, and G.~Yang, ``Towards reliable and explainable ai model for pulmonary nodule diagnosis,'' {\em Biomedical Signal Processing and Control}, vol.~88, p.~105646, 2024.

\bibitem{gonzalez2023lifelong}
C.~Gonz{\'a}lez, A.~Ranem, D.~Pinto~dos Santos, A.~Othman, and A.~Mukhopadhyay, ``Lifelong nnu-net: a framework for standardized medical continual learning,'' {\em Scientific Reports}, vol.~13, no.~1, p.~9381, 2023.

\bibitem{zhao2022efficient}
Y.~Zhao, C.~Yang, A.~Schweidtmann, and Q.~Tao, ``Efficient bayesian uncertainty estimation for nnu-net,'' in {\em International Conference on Medical Image Computing and Computer-Assisted Intervention}, pp.~535--544, Springer, 2022.

\bibitem{joshi2024leveraging}
S.~Joshi, R.~Osuala, L.~Garrucho, A.~Tsirikoglou, J.~del Riego, K.~Gwo{\'z}dziewicz, K.~Kushibar, O.~Diaz, and K.~Lekadir, ``Leveraging epistemic uncertainty to improve tumour segmentation in breast mri: an exploratory analysis,'' in {\em Medical Imaging 2024: Image Processing}, vol.~12926, pp.~292--300, SPIE, 2024.

\bibitem{vorberg2024bayesian}
L.~Vorberg, H.~Ditt, M.~S{\"u}hling, A.~Maier, N.~Murray, S.~Nicolaou, and O.~Taubmann, ``Bayesian uncertainty estimation improves nnu-net generalization to unseen sites for stroke lesion segmentation,'' in {\em MICCAI Challenge on Ischemic Stroke Lesion Segmentation}, pp.~22--30, Springer, 2024.

\bibitem{Koutsoubis2024}
N.~Koutsoubis, Y.~Yilmaz, R.~P. Ramachandran, S.~Matthew, and R.~Ghulam, ``Privacy preserving federated learning in medical imaging with uncertainty estimation.'' arXiv: 2406.12815, 6 2024.

\bibitem{sandhu2023medical}
S.~S. Sandhu, H.~T. Gorji, P.~Tavakolian, K.~Tavakolian, and A.~Akhbardeh, ``Medical imaging applications of federated learning,'' {\em Diagnostics}, vol.~13, no.~19, p.~3140, 2023.

\bibitem{teo2024federated}
Z.~L. Teo, L.~Jin, N.~Liu, S.~Li, D.~Miao, X.~Zhang, W.~Y. Ng, T.~F. Tan, D.~M. Lee, K.~J. Chua, {\em et~al.}, ``Federated machine learning in healthcare: A systematic review on clinical applications and technical architecture,'' {\em Cell Reports Medicine}, vol.~5, no.~2, 2024.

\bibitem{Che_FedEvi_MICCAI2024}
J.~Chen, B.~Ma, H.~Cui, and Y.~Xia, ``{ FedEvi: Improving Federated Medical Image Segmentation via Evidential Weight Aggregation },'' in {\em proceedings of Medical Image Computing and Computer Assisted Intervention -- MICCAI 2024}, vol.~LNCS 15010, Springer Nature Switzerland, October 2024.

\bibitem{schutte2024fedgs}
P.~Schutte, V.~Corbetta, R.~Beets-Tan, and W.~Silva, ``Fedgs: Federated gradient scaling for heterogeneous medical image segmentation,'' in {\em International Conference on Medical Image Computing and Computer-Assisted Intervention}, pp.~246--255, Springer, 2024.

\bibitem{Xu2023FedMENU}
X.~Xu, H.~H. Deng, J.~Gateno, and P.~Yan, ``Federated multi-organ segmentation with inconsistent labels,'' {\em IEEE Transactions on Medical Imaging}, vol.~42, no.~10, pp.~2948--2960, 2023.

\bibitem{vaswani2017attention}
A.~Vaswani, N.~Shazeer, N.~Parmar, J.~Uszkoreit, L.~Jones, A.~N. Gomez, {\L}.~Kaiser, and I.~Polosukhin, ``Attention is all you need,'' {\em Advances in neural information processing systems}, vol.~30, 2017.

\bibitem{shiri2023multi}
I.~Shiri, B.~Razeghi, A.~V. Sadr, M.~Amini, Y.~Salimi, S.~Ferdowsi, P.~Boor, D.~G{\"u}nd{\"u}z, S.~Voloshynovskiy, and H.~Zaidi, ``Multi-institutional pet/ct image segmentation using federated deep transformer learning,'' {\em Computer Methods and Programs in Biomedicine}, vol.~240, p.~107706, 2023.

\bibitem{liu2024fedfmsexploringfederatedfoundation}
Y.~Liu, G.~Luo, and Y.~Zhu, ``Fedfms: Exploring federated foundation models for medical image segmentation,'' 2024.

\bibitem{kirillov2023segment}
A.~Kirillov, E.~Mintun, N.~Ravi, H.~Mao, C.~Rolland, L.~Gustafson, T.~Xiao, S.~Whitehead, A.~C. Berg, W.-Y. Lo, P.~Dollár, and R.~Girshick, ``Segment anything,'' 2023.

\bibitem{MedSAM}
J.~Ma, Y.~He, F.~Li, L.~Han, C.~You, and B.~Wang, ``Segment anything in medical images,'' {\em Nature Communications}, vol.~15, p.~654, 2024.

\bibitem{Enmao21}
E.~Diao, J.~Ding, and V.~Tarokh, ``Heterofl: Computation and communication efficient federated learning for heterogeneous clients.'' arXiv: 2010.01264, 2021.

\bibitem{horvath2021fjord}
S.~Horvath, S.~Laskaridis, M.~Almeida, I.~Leontiadis, S.~Venieris, and N.~Lane, ``Fjord: Fair and accurate federated learning under heterogeneous targets with ordered dropout,'' {\em Advances in Neural Information Processing Systems}, vol.~34, pp.~12876--12889, 2021.

\bibitem{alam2022fedrolex}
S.~Alam, L.~Liu, M.~Yan, and M.~Zhang, ``Fedrolex: Model-heterogeneous federated learning with rolling sub-model extraction,'' {\em Advances in neural information processing systems}, vol.~35, pp.~29677--29690, 2022.

\bibitem{flower2020}
D.~J. Beutel {\em et~al.}, ``Flower: A friendly federated learning research framework,'' {\em arXiv: 2007.14390}, 2020.

\bibitem{Li2018}
T.~Li, A.~K. Sahu, M.~Zaheer, M.~Sanjabi, A.~Talwalkar, and V.~Smith, ``Federated optimization in heterogeneous networks.'' arXiv: 1812.06127, 2020.

\bibitem{Garrucho2024}
L.~Garrucho, K.~Kushibar, C.-A. Reidel, S.~Joshi, R.~Osuala, A.~Tsirikoglou, M.~Bobowicz, J.~Del~Riego, A.~Catanese, K.~Gwo{\'z}dziewicz, {\em et~al.}, ``A large-scale multicenter breast cancer dce-mri benchmark dataset with expert segmentations,'' {\em Scientific data}, vol.~12, no.~1, p.~453, 2025.

\bibitem{eucanimage}
{EuCanImage Project}, ``Towards a european cancer imaging platform for enhanced artificial intelligence in oncology..'' \url{https://eucanimage.eu/}.
\newblock Accessed July 3, 2025.

\bibitem{Campello2021}
V.~M. Campello {\em et~al.}, ``Multi-centre, multi-vendor and multi-disease cardiac segmentation: The m\&ms challenge,'' {\em IEEE Transactions on Medical Imaging}, vol.~40, no.~12, pp.~3543--3554, 2021.

\bibitem{Isla2023}
C.~Martín-Isla {\em et~al.}, ``Deep learning segmentation of the right ventricle in cardiac mri: The m\&ms challenge,'' {\em IEEE Journal of Biomedical and Health Informatics}, vol.~27, no.~7, pp.~3302--3313, 2023.

\bibitem{aimix}
{AIMIX Project}, ``Inclusive artificial intelligence for accessible medical imaging across resource-limited settings.'' \url{https://aimix-erc.eu/}.
\newblock Accessed July 3, 2025.

\bibitem{Carneiro2008}
G.~Carneiro, B.~Georgescu, S.~Good, and D.~Comaniciu, ``Detection and measurement of fetal anatomies from ultrasound images using a constrained probabilistic boosting tree,'' {\em IEEE Transactions on Medical Imaging}, vol.~27, no.~9, pp.~1342--1355, 2008.

\bibitem{sappia_2024_12697994}
M.~S. Sappia, C.~L. de~Korte, B.~van Ginneken, D.~Ninalga, S.~Kondo, S.~Kasai, K.~Hirasawa, T.~Akumu, C.~Mart{\'\i}n-Isla, K.~Lekadir, {\em et~al.}, ``Acouslic-ai challenge report: Fetal abdominal circumference measurement on blind-sweep ultrasound data from low-income countries,'' {\em Medical Image Analysis}, p.~103640, 2025.

\bibitem{UNETR}
A.~Hatamizadeh {\em et~al.}, ``{ UNETR: Transformers for 3D Medical Image Segmentation },'' in {\em 2022 IEEE/CVF Winter Conference on Applications of Computer Vision (WACV)}, (Los Alamitos, CA, USA), pp.~1748--1758, IEEE Computer Society, Jan. 2022.

\bibitem{nnformer}
H.-Y. Zhou {\em et~al.}, ``nnformer: Volumetric medical image segmentation via a 3d transformer,'' {\em IEEE Transactions on Image Processing}, vol.~32, pp.~4036--4045, 2023.

\bibitem{lekadir2025future}
K.~Lekadir, A.~F. Frangi, A.~R. Porras, B.~Glocker, C.~Cintas, C.~P. Langlotz, E.~Weicken, F.~W. Asselbergs, F.~Prior, G.~S. Collins, {\em et~al.}, ``Future-ai: International consensus guideline for trustworthy and deployable artificial intelligence in healthcare,'' {\em bmj}, vol.~388, 2025.

\bibitem{fabila2025federated}
J.~Fabila, L.~Garrucho, V.~M. Campello, C.~Mart{\'\i}n-Isla, and K.~Lekadir, ``Federated learning in low-resource settings: A chest imaging study in africa--challenges and lessons learned,'' {\em arXiv preprint arXiv:2505.14217}, 2025.

\bibitem{geiping2020inverting}
J.~Geiping, H.~Bauermeister, H.~Dr{\"o}ge, and M.~Moeller, ``Inverting gradients-how easy is it to break privacy in federated learning?,'' {\em Advances in neural information processing systems}, vol.~33, pp.~16937--16947, 2020.

\bibitem{nasr2019comprehensive}
M.~Nasr, R.~Shokri, and A.~Houmansadr, ``Comprehensive privacy analysis of deep learning: Passive and active white-box inference attacks against centralized and federated learning,'' in {\em 2019 IEEE symposium on security and privacy (SP)}, pp.~739--753, IEEE, 2019.

\bibitem{dp_n_federated_learning}
A.~E. Ouadrhiri and A.~Abdelhadi, ``Differential privacy for deep and federated learning: A survey,'' {\em IEEE Access}, vol.~10, pp.~22359--22380, 2022.

\bibitem{kaissis2020secure}
G.~A. Kaissis, M.~R. Makowski, D.~R{\"u}ckert, and R.~F. Braren, ``Secure, privacy-preserving and federated machine learning in medical imaging,'' {\em Nature Machine Intelligence}, vol.~2, no.~6, pp.~305--311, 2020.

\bibitem{ziller2021differentially}
A.~Ziller, D.~Usynin, N.~Remerscheid, M.~Knolle, M.~Makowski, R.~Braren, D.~Rueckert, and G.~Kaissis, ``Differentially private federated deep learning for multi-site medical image segmentation,'' {\em arXiv preprint arXiv:2107.02586}, 2021.

\bibitem{riess2024complex}
A.~Riess, A.~Ziller, S.~Kolek, D.~Rueckert, J.~Schnabel, and G.~Kaissis, ``Complex-valued federated learning with differential privacy and mri applications,'' in {\em International Conference on Medical Image Computing and Computer-Assisted Intervention}, pp.~191--203, Springer, 2024.

\bibitem{raggio2025fedsynthct}
C.~B. Raggio, M.~K. Zabaleta, N.~Skupien, O.~Blanck, F.~Cicone, G.~L. Cascini, P.~Zaffino, L.~Migliorelli, and M.~F. Spadea, ``Fedsynthct-brain: A federated learning framework for multi-institutional brain mri-to-ct synthesis,'' {\em Computers in Biology and Medicine}, vol.~192, p.~110160, 2025.

\bibitem{osuala2023data}
R.~Osuala, K.~Kushibar, L.~Garrucho, A.~Linardos, Z.~Szafranowska, S.~Klein, B.~Glocker, O.~Diaz, and K.~Lekadir, ``Data synthesis and adversarial networks: A review and meta-analysis in cancer imaging,'' {\em Medical Image Analysis}, vol.~84, p.~102704, 2023.

\bibitem{osuala2024enhancing}
R.~Osuala, D.~M. Lang, A.~Riess, G.~Kaissis, Z.~Szafranowska, G.~Skorupko, O.~Diaz, J.~A. Schnabel, and K.~Lekadir, ``Enhancing the utility of privacy-preserving cancer classification using synthetic data,'' in {\em Deep Breast Workshop on AI and Imaging for Diagnostic and Treatment Challenges in Breast Care}, pp.~54--64, Springer, 2024.

\bibitem{zhou2024secure}
I.~Zhou, F.~Tofigh, M.~Piccardi, M.~Abolhasan, D.~Franklin, and J.~Lipman, ``Secure multi-party computation for machine learning: A survey,'' {\em IEEE Access}, 2024.

\end{thebibliography}

\end{document}